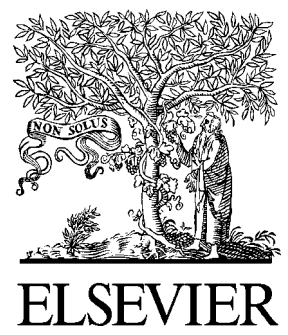

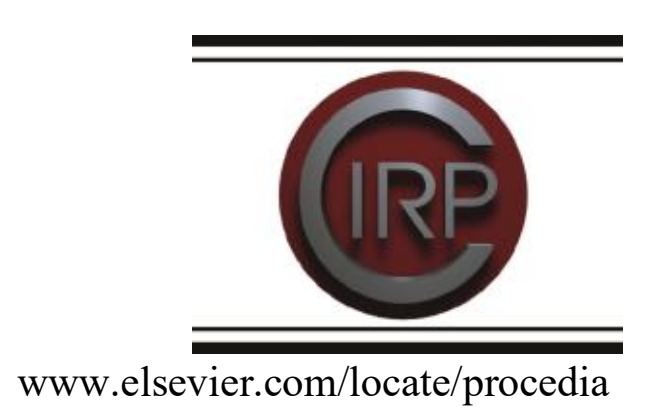

# APPROVE: Visual End-User-in-the-Loop Robot Programming with LLMs

Bijan Kavousian[a]*, Miray Özakkas[b], Josefine Monnet[a], Oliver Petrovic[a], Christian Brecher[a]

[a]Laboratory for Machine Tools and Production Engineering WZL of RWTH Aachen University, Steinbachstraße 19, 52074 Aachen, Germany
[b]RWTH Aachen University, Templergraben 55, 52062 Aachen, Germany

* Corresponding author. Tel.: +49 241 80-25423; *E-mail address:* b.kavousian@wzl.rwth-aachen.de

**Abstract**

Programming robots remains challenging for non-experts, as traditional methods require expert knowledge and even block-based interfaces often lack flexibility. Recent work has explored Large Language Models (LLMs) to automatically generate robot programs from natural language, but these systems remain limited by a lack of transparency, missing mechanisms to ensure alignment with user intent, and little support for reuse. We present APPROVE (AI-Powered Programming for Robots with Visual End-User Feedback), an LLM-based multi-modal end-user programming framework that integrates natural language input with a block-based interface and an explicit user confirmation step. Generated programs are visualized using a block-based interface in Blockly, allowing users to confirm, modify, or reject them before execution. Confirmed functions are stored in a library for reuse, gradually building a set of reliable program components. Our approach contributes a human-centered design for LLM-based robot programming that emphasizes user trust, intent alignment, and reusability.

## 1. Introduction

Robots are increasingly deployed across industry and research [1]. In high-mix, low-volume production environments, where product variants and task setups change frequently, programming and reconfiguration often become the primary bottleneck: Robotics experts are scarce, and domain experts on the shop floor are required to adapt robot behavior themselves [2]. This creates a need for end-user programming that is intuitive yet reliable, offering transparency, user control, and reusability.

Several approaches have been proposed to simplify robot programming, including block-based interfaces, skill-based programming, LLM-based code generation, and vision-language-action (VLA) models. However, these approaches typically trade off transparency and control against flexibility and generalization, rarely providing both.

To address this gap, we present APPROVE, an end-user robot programming framework designed to make robot code creation transparent, controllable, and reusable. The user interface is shown in Figure 1. Rather than hiding code generation, APPROVE combines natural-language specification with a visual representation that users can inspect at a glance. An explicit confirmation step keeps users in charge before execution, while lightweight edits, either as visual changes or brief language instructions, support quick correction. Confirmed functions are provided as reusable building blocks that accelerate development, improve reproducibility and safety, and build trust in the system's behavior.

This paper contributes (i) a language-to-code pipeline with a visual confirmation layer, (ii) an interaction design that centers explicit user control to improve transparency and intent alignment, and (iii) a function library mechanism based on skill-based programming that promotes reuse and trust by accumulating confirmed program components.

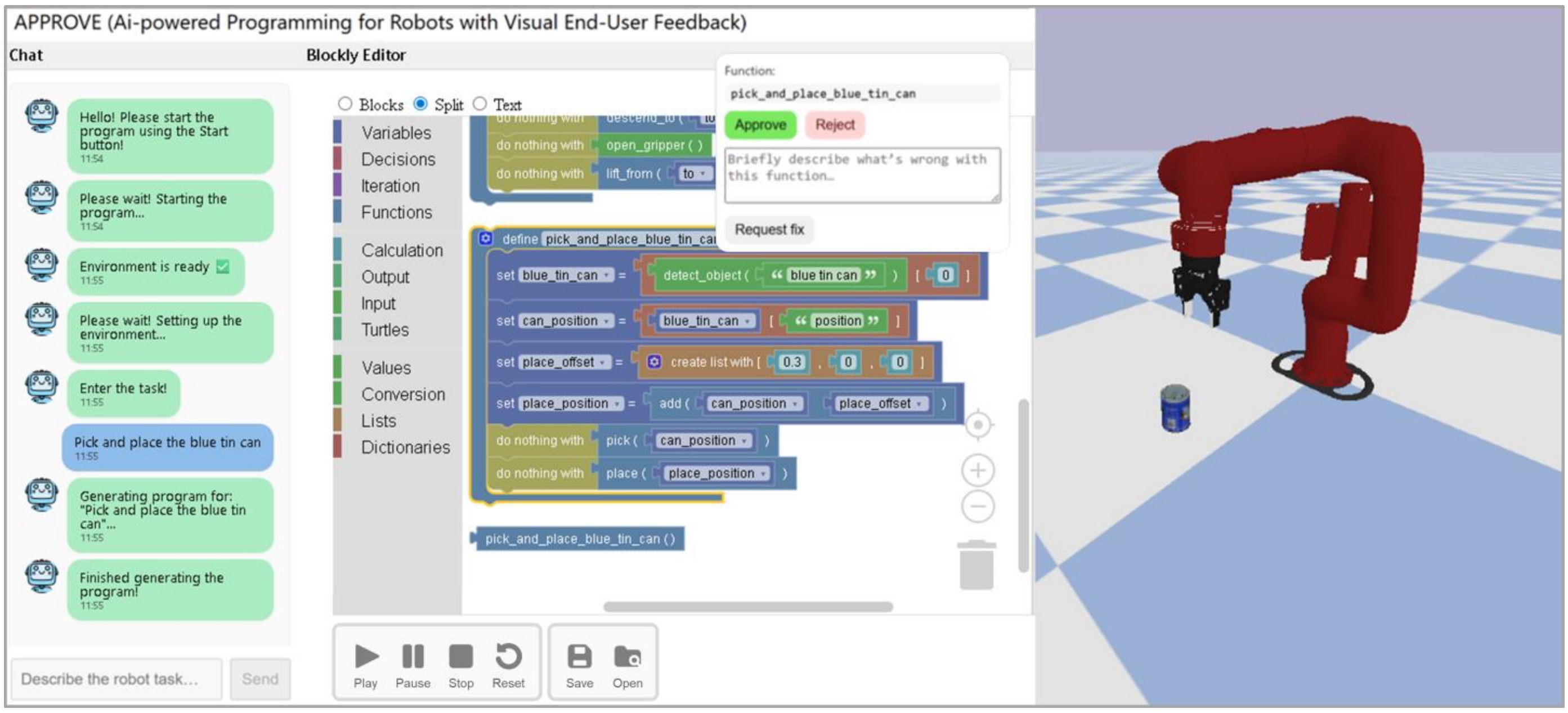

Figure 1: The APPROVE interface combines natural language input and block-based visualization with user confirmation and simulation of robot behavior.

## 2. Related Work

Traditional robot programming provides fine-grained control but remains inaccessible to non-experts [2]. Natural-language interfaces lower the entry barrier, yet suffer from ambiguity and opacity; users often cannot see what the system understood, verify the generated program, or reuse it. These limitations restrict transparency, reproducibility, and trust.

An emerging path towards robot applications without explicit programming are Vision-Language-Action (VLA) models. Instead of offering explicit programming interfaces, VLAs such as RT-1 [3], RT-2 [4], OpenVLA [5] or Gemini Robotics [6] directly map multimodal inputs–typically visual observations and natural-language instructions–to robot actions. This allows robots to generalize beyond pre-programmed tasks and even execute instructions for previously unseen scenarios [7]. However, current performance remains limited: Even for State-of-the-Art-Models, average reported success rates vary widely across benchmarks, ranging from around 20 % in challenging zero-shot tasks to above 80 % in constrained scenarios [5, 6]. While these results are impressive, in most cases they are not suitable for real-world use yet. Another difficulty with VLA models is their black-box behavior, which limits transparency and makes it difficult for users to understand, verify, or correct robot behavior.

These limitations underline the complementary role of VLAs: while they showcase the potential of foundation models for robotics, more controllable and human-centered approaches for intuitive programming remain essential.

One widely adopted approach are block-based programming languages, which are often perceived as more intuitive than their text-based counterparts [8]. This is a reason why many robot programming applications rely on block-based programming. Weintrop et al. [9] demonstrate that block-based methods can be effectively applied to robot applications while maintaining accessibility. Furthermore, most cobots use block-based interfaces as their built-in programming method [10].

Beyond block-based methods, skill-based robot programming provides a structured hierarchy of primitives, skills, and tasks [11]. Primitives capture device-level actions such as sensing or actuation, while skills combine them into meaningful building blocks like “pick” or “move.” On this basis, even users with limited robotics expertise can compose complete tasks by arranging existing skills [12].

A prominent recent direction is the integration of Large Language Models (LLMs) into robot programming. These approaches map language to executable robot programs and refine them through dialogue (e.g., Alchemist [13]). They enable flexible task specification and have been applied to retrieve production data and select robot actions [14], to generate robot programs including trajectories by iteratively calling functions and re-processing the output [15], or to generate entire programs end-to-end. Applications range from laboratory environments [13] to service robotics [16, 17]. Other approaches use LLMs to generate Behavior Trees for robots [18] or propose hierarchical strategies that structure programming from coarse to fine and iteratively refine it with human feedback, while still relying on code-only outputs. [19]. However, most approaches still treat code generation as a largely hidden process and do not require explicit user confirmation before execution [13, 20, 21]. As a result, mechanisms for inspection, correction, and library-driven reuse remain limited, leaving intent alignment and safety concerns insufficiently addressed.

More recent approaches combine block-based interfaces with chat interfaces. CAPIRCI combined a chat interface with a block editor for realizing pick-and-place tasks. Their results suggest that presenting the generated code as blocks and allowing modifications increases transparency and trust [22]. Subsequent iterations extended the system with LLM-based natural language processing [23] and later accelerated the interaction by enabling the direct generation of initial program suggestions instead of requiring long chat sequences [21].

Table 1. Comparison of approaches for intuitive robot programming.

| Approach | Principle | Example | Pros | Cons |
|---|---|---|---|---|
| Block-based | Visual blocks | [8, 9] | Transparent | Limited flexibility |
| Skill-based | Hierarchical skills | [11, 12] | Modular, reusable | Often not automated |
| VLA | End-to-end policy | [3–6] | Generalization | Black-Box |
| LLM-based | NL → code | [13–23] | Intuitive, flexible | Opaque |

Table 1 summarizes the key characteristics of existing approaches and highlights that they either emphasize transparency and structured programming or flexibility and generalization, but rarely both. Systems that combine flexible task specification with transparency and explicit user control remain limited. We address this gap with APPROVE, combining natural language interaction with explicit visual verification and reusable program structures.

## 3. System & Workflow Overview

This section explains APPROVE by walking through the user workflow and highlighting the system components involved at each step.

The overall structure is given in Figure 2. Natural Language instructions from the human end user are processed using an LLM. This allows leveraging the strengths of LLMs: Natural language can be understood, enabling people without deep robotics knowledge to express their intentions for the robot program without needing to adhere to a specific format [24]. Using natural language can make interaction easier [25]. Furthermore, as discussed in Section 1, LLMs are already able to generate reasonably good robotics code in many cases. By combining these capabilities, the system generates initial program suggestions that the user can then verify and refine through the GUI.

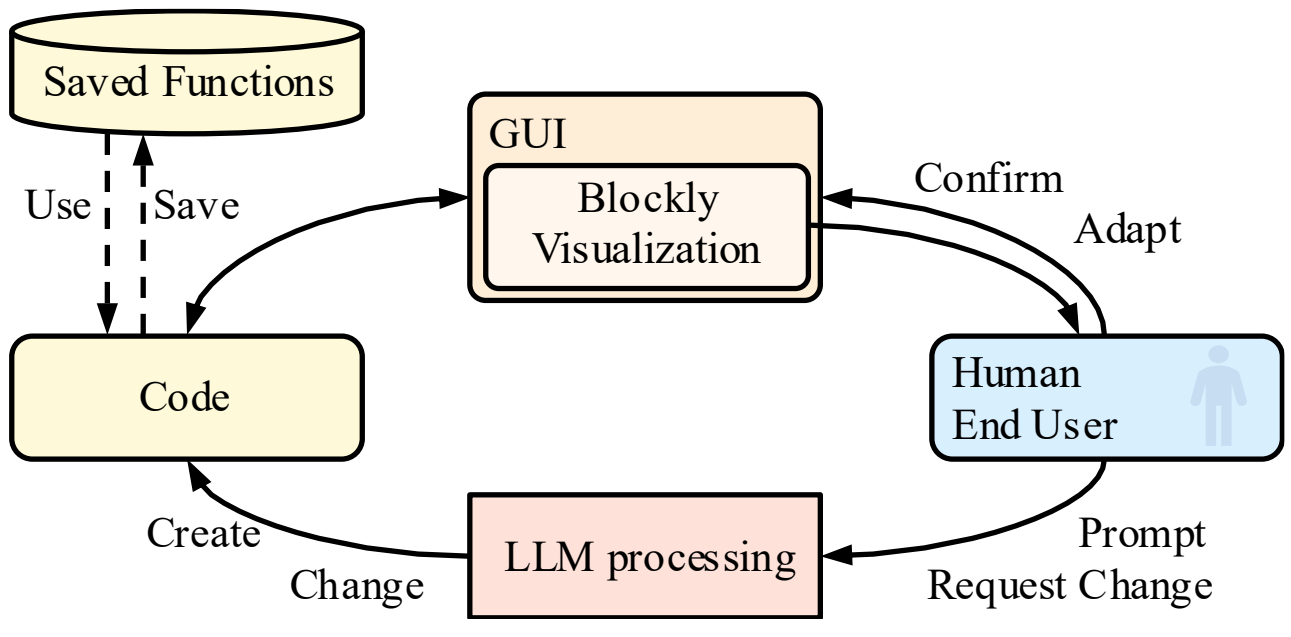

Figure 2: Overview of the system. The human end user can interact with the code through the LLM and the GUI including its Blockly visualization.

After the user states their request via text, the system uses the OpenAI GPT-5 API [26] to generate python code.

This offers several advantages over directly generating blocks: (i) LLMs are trained on large amounts of Python code, making it easier for them to produce Python code rather than specialized Blockly JSONs or similar [27], (ii) new primitive functions can easily be incorporated, since they are usually available in python, and (iii) BlockMirror also allows viewing the Python code [28], which some users may prefer.

The GUI then provides a Blockly-based visualization [29] of the generated code and a 3D simulation of the robotic system for the user to review the suggested program using BlockMirror [28]. The Blockly interface provides an at-a-glance view of the program's logic by showing top-level functions first. From there, users can either confirm the suggested function, edit it, or request an edit from the LLM:

*Confirm:* If the user agrees with the LLM on the suggested function, they can confirm it. The confirmation mechanism follows a bottom-up approach: A higher-level function can only be confirmed once the functions it calls are confirmed. It is then marked visually and stored in the function library for future use.

*Manually edit:* Users can directly adjust the Blockly code, for example by reordering functions, modifying parameters, or inserting additional blocks from the library.

*Request edit:* Alternatively, users can ask the LLM to revise the program. The current state of the code, including any manual changes, is provided together with the user's instruction. The LLM returns an updated version, and another iteration of user checks is started.

Confirmed functions are stored in the function library. This library is provided to the LLM as part of the context, enabling it to reuse previously confirmed functions instead of regenerating them. This reduces the likelihood of errors and simplifies verification, as users can stop inspection at the level of confirmed functions. Code generation is constrained to a predefined whitelist of callable functions and permitted Python built-ins, thereby enforcing a bounded and executable action space.

The generated program is organized as a hierarchy of functions according to the abstraction layers in skill-based robot programming as shown in Figure 3 [11]. At the lowest level are primitives, which provide basic device-specific capabilities such as actuation or sensing. These are complemented by services, which represent hardware-independent base operations, such as object detection or trajectory generation. Together, primitives and services are combined into skills, representing reusable building blocks that encapsulate meaningful actions such as picking, moving, or inspecting. At the highest level, tasks define the overall execution sequence by orchestrating multiple skills to achieve a complete operation. Because all layers share a uniform representation in the system, even complex tasks can themselves be reused as subtasks within larger behaviors, supporting modularity and scalability.

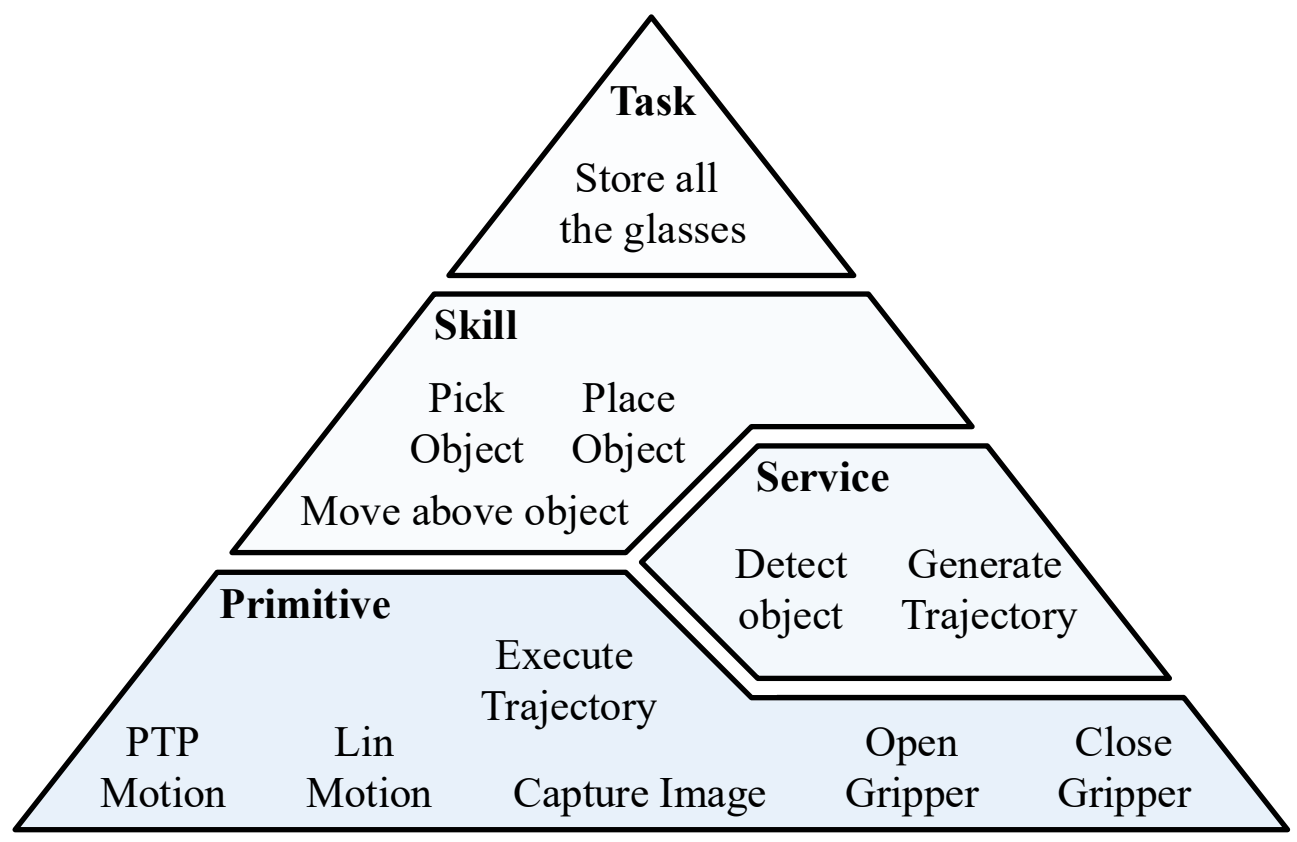

Figure 3: Hierarchy of abstraction levels in skill-based robot programming, based on the structure proposed in [14], with example functions adapted to our framework.

In the current prototype, the primitive layer comprises motion, gripper, and perception primitives. Motion primitives include move_to(pose, speed, frame) and get_current_pose(), which provide access to the robot's Cartesian state and enable the construction of higher-level behaviors. For example, relative motions are not implemented as built-in primitives but can be derived by combining get_current_pose() with move_to(...); once confirmed, such derived functions become reusable elements in the library. Gripper primitives such as open_gripper() and close_gripper(force) provide control over the end-effector.

The perception service get_object_pose(label) provides a vision module based on Grounding DINO [30], a text-conditioned object detection model. The model processes a natural-language object description and returns bounding boxes, which are combined with depth data to estimate the object's 3D pose in the robot's coordinate system. Together,

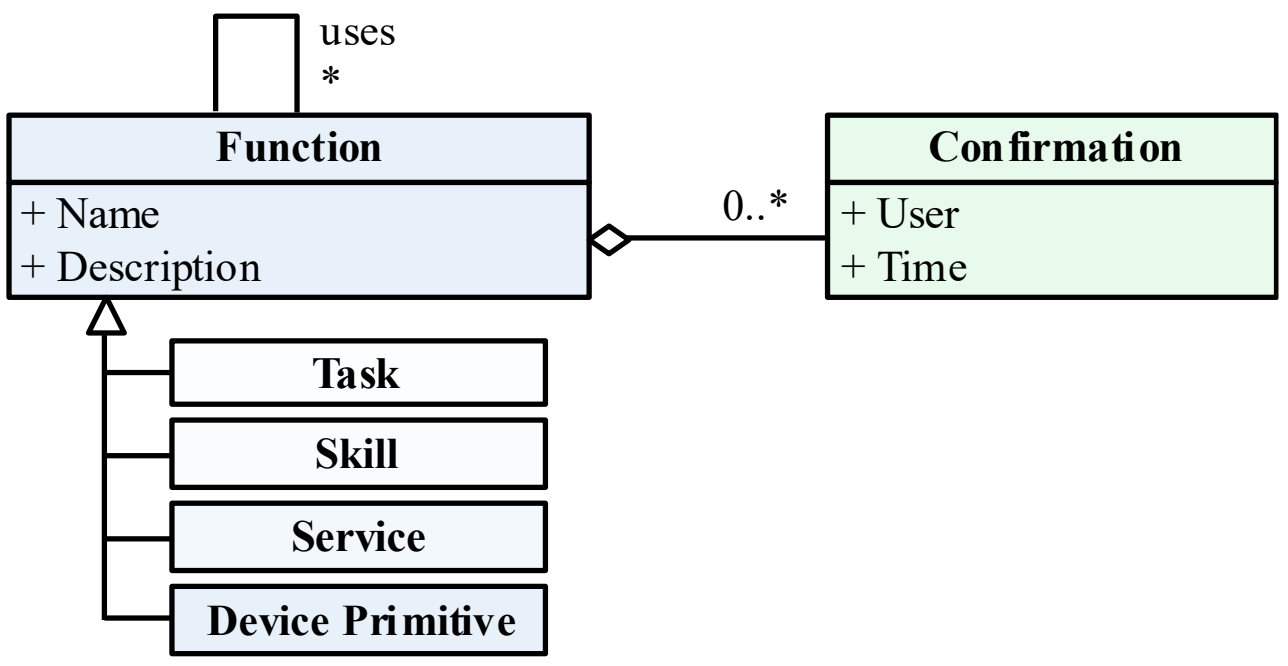

Figure 4: Model of functions and confirmations in APPROVE. Functions provide a uniform representation across abstraction levels from device primitives to tasks. User confirmations are modeled separately and attached to functions to record responsibility and time

these primitives and services define the minimal executable action space to which code generation is restricted.

To promote reuse and trust, each function in the function library carries the same properties (see Figure 4): a clear name and description that state intent and expected effects, a parameter signature that enables reuse across contexts without rewriting, and user confirmations recorded with user and timestamp. Confirmed functions are stored in the function library, displayed with a confirmation badge in the GUI, and made available for subsequent tasks.

## 4. Formative User Insights

To gain initial insights into the usability of the system, we conducted a small pilot study. The study aimed to assess the current state of the prototype and identify opportunities for improvement. Three participants (2 male, 1 female; ages 24–32, M = 27, SD = 4.2), none of whom were involved in the project, were invited to use the system after a short introduction to its interface and controls. All participants provided informed consent prior to the study. Their task was to program a simple pick-and-place operation and to share feedback on their experience, allowing us to observe how they interacted with the interface and where difficulties or strengths of the approach emerged. All participants were eventually able to successfully program the task using the system, i.e., the task was successfully executed in simulation.

With respect to the interaction flow and transparency, one participant preferred to run the visualization of the program before approving it. While the intended workflow was to first inspect the code, this alternative mode of use is already supported by the system and should be considered in future experiments. Another participant suggested that the system should ask clarifying questions in ambiguous cases, which could help align the program more quickly with the user's intent. A participant also emphasized the value of a before–after comparison when the LLM revises code based on user input, as such a feature would make changes more transparent and help users understand how their feedback influences the generated program.

Finally, participants highlighted several strengths of the approach. The reuse of functions was perceived as particularly helpful: when programming a second, similar task, inspection was easier because previously confirmed skills were already available. This indicates that the system becomes more usable over time as the function library grows. In addition, the dual presentation of programs in both block-based and text-based form was described as clear and understandable.

In summary, the pilot study suggests that the system is usable in principle, as all participants were able to complete the task successfully. Since it was a small pilot with only three participants, the findings are limited and should be seen as first indications rather than general evidence. At the same time, the feedback pointed to clear opportunities for improvement, which will guide the next steps in refining the system.

## 5. Discussion

With LLMs becoming increasingly successful at generating robot programs, and considering that the target users are non-experts in robotics, an important design question arises: why should the confirmation step be assigned to the user rather than to the system itself? While an LLM can check whether a program is logically consistent, it cannot ensure that the program matches the user's intent. By requiring explicit user confirmation, potential inaccuracies or ambiguities in prompts can be compensated, ensuring that the robot's behavior reflects the user's intended task.

Another important factor concerns the user's perception of the system. Letting the user explicitly confirm the program promotes transparency and predictability, both of which are known to support trust [31]. Moreover, leaving control to the user can itself positively affect trust [32]. Another relevant perspective is the concept of a formal work agreement, in which explicit responsibilities enable collaboration even under conditions of limited trust [33]. This effect is to be evaluated in a larger user study.

## 6. Conclusion

This paper presented APPROVE, a framework for end-user robot programming that integrates LLM-based code generation with visual representation and explicit human confirmation. The approach combines natural language input with Blockly visualization, enabling users to inspect, adapt, and approve programs before execution. Confirmed functions are stored in a library, promoting reuse and gradually building a corpus of trustworthy components.

By aligning powerful generative models with human oversight, the system addresses key challenges of transparency, control, and trust in end-user programming. The prototype implementation demonstrates that such an approach can make robot programming more accessible for non-experts while still supporting reproducibility in dynamic environments.

Overall, APPROVE illustrates a step towards more transparency, control, and reusability in LLM-assisted robot programming. The results highlight the potential of combining LLMs with explicit human feedback to create systems that are not only usable but also trustworthy in practice.

### *6.1. Future Work*

To further advance this vision, future work will extend evaluation with a larger and more diverse group of participants to better understand usability across expertise levels and identify opportunities for improving the interface. On the technical side, the present implementation offers a small but functional set of primitives; expanding this library with more complex and domain-specific primitives and skills will broaden the range of tasks while preserving reusability and trust. As the function library grows, Retrieval-Augmented Generation [34] could be incorporated to access the functions more efficiently. Finally, connecting the system to physical robots will enable us to test whether generated programs transfer reliably from simulation to hardware and whether the human-in-the-loop mechanism remains effective under real-world conditions.

## Acknowledgements

The studies described in this paper were conducted as part of the research and development project "AKzentE4.0". This research and development project is funded by the German Federal Ministry of Research, Technology and Space (BMFTR) under the funding measure "Future of Work: Regional Competence Centers of Labor Research. Designing new forms of work through artificial intelligence" in the program "Innovations for tomorrow's production, services and work" (funding code: 02L19C400) and supervised by the Project Management Agency Karlsruhe (PTKA).